\documentclass{article}

\usepackage{microtype}
\usepackage{graphicx}
\usepackage{subfigure}
\usepackage{booktabs} 

\usepackage{hyperref}
\usepackage{multirow}
\usepackage{array}
\usepackage[dvipsnames, table]{xcolor}

\makeatletter

\newcommand{\Rmnum}[1]{\expandafter\@slowromancap\romannumeral#1@}
\makeatletter

\usepackage[accepted]{icml2024}

\usepackage{amsmath}
\usepackage{amssymb}
\usepackage{mathtools}
\usepackage{amsthm}

\usepackage[capitalize,noabbrev]{cleveref}

\theoremstyle{plain}

\theoremstyle{definition}

\theoremstyle{remark}

\usepackage[textsize=tiny]{todonotes}

\icmltitlerunning{DSD-DA: Distillation-based Source Debiasing for Domain Adaptive Object Detection}

\begin{document}

\twocolumn[
\icmltitle{DSD-DA: Distillation-based Source Debiasing for \\ Domain Adaptive Object Detection}



\icmlsetsymbol{equal}{*}

\begin{icmlauthorlist}
\icmlauthor{Yongchao Feng}{buaa}
\icmlauthor{Shiwei Li}{hyy}
\icmlauthor{Yingjie Gao}{buaa}
\icmlauthor{Ziyue Huang}{buaa}
\icmlauthor{Yanan Zhang}{buaa}
\icmlauthor{Qingjie Liu}{buaa,zgc,hyy}
\icmlauthor{Yunhong Wang}{buaa,hyy}

\end{icmlauthorlist}

\icmlaffiliation{buaa}{State Key Laboratory of Virtual Reality Technology and Systems, Beihang University, Beijing, China.}
\icmlaffiliation{zgc}{Zhongguancun Laboratory, Beijing, China}
\icmlaffiliation{hyy}{Hangzhou Innovation Institute, Beihang University, Hangzhou, China.}


\icmlcorrespondingauthor{Qingjie Liu}{qingjie.liu@buaa.edu.cn}

\icmlkeywords{Machine Learning, ICML}

\vskip 0.3in
]



\printAffiliationsAndNotice{}  

\begin{abstract}
\vspace{-1pt}
 Though feature-alignment based Domain Adaptive Object Detection (DAOD) methods have achieved remarkable progress, they ignore the source bias issue, \textit{i.e.,} the detector tends to acquire more source-specific knowledge, impeding its generalization capabilities in the target domain. Furthermore, these methods face a more formidable challenge in achieving consistent classification and localization in the target domain compared to the source domain.
 To overcome these challenges, we propose a novel Distillation-based Source Debiasing (DSD) framework for DAOD, which can distill domain-agnostic knowledge from a pre-trained teacher model, improving the detector's performance on both domains.
 In addition, we design a Target-Relevant Object Localization Network (TROLN), which can mine target-related localization information from source and target-style mixed data. Accordingly, we present a Domain-aware Consistency Enhancing (DCE) strategy, in which these information are formulated into a new localization representation to further refine classification scores in the testing stage, achieving a harmonization between classification and localization. Extensive experiments have been conducted to manifest the effectiveness of this method, which consistently improves the strong baseline by large margins, outperforming existing alignment-based works.
\end{abstract}

\section{Introduction}
\label{submission}

State-of-the-art object detectors \cite{redmon2018yolov3,ren2015faster,tian2019fcos} have demonstrated impressive performance when the training and testing data exhibit consistent distributions.
However, their performance diminishes drastically when applied to novel domains, primarily due to domain shift \cite{chen2018domain}, which impedes the generalization and transferability of the detectors across different scenes. 
The inability of object detectors to adapt to novel domains hampers their practical applicability in real-world scenarios. 

\begin{figure}[t]
\centering
\includegraphics[width=1.00\columnwidth]{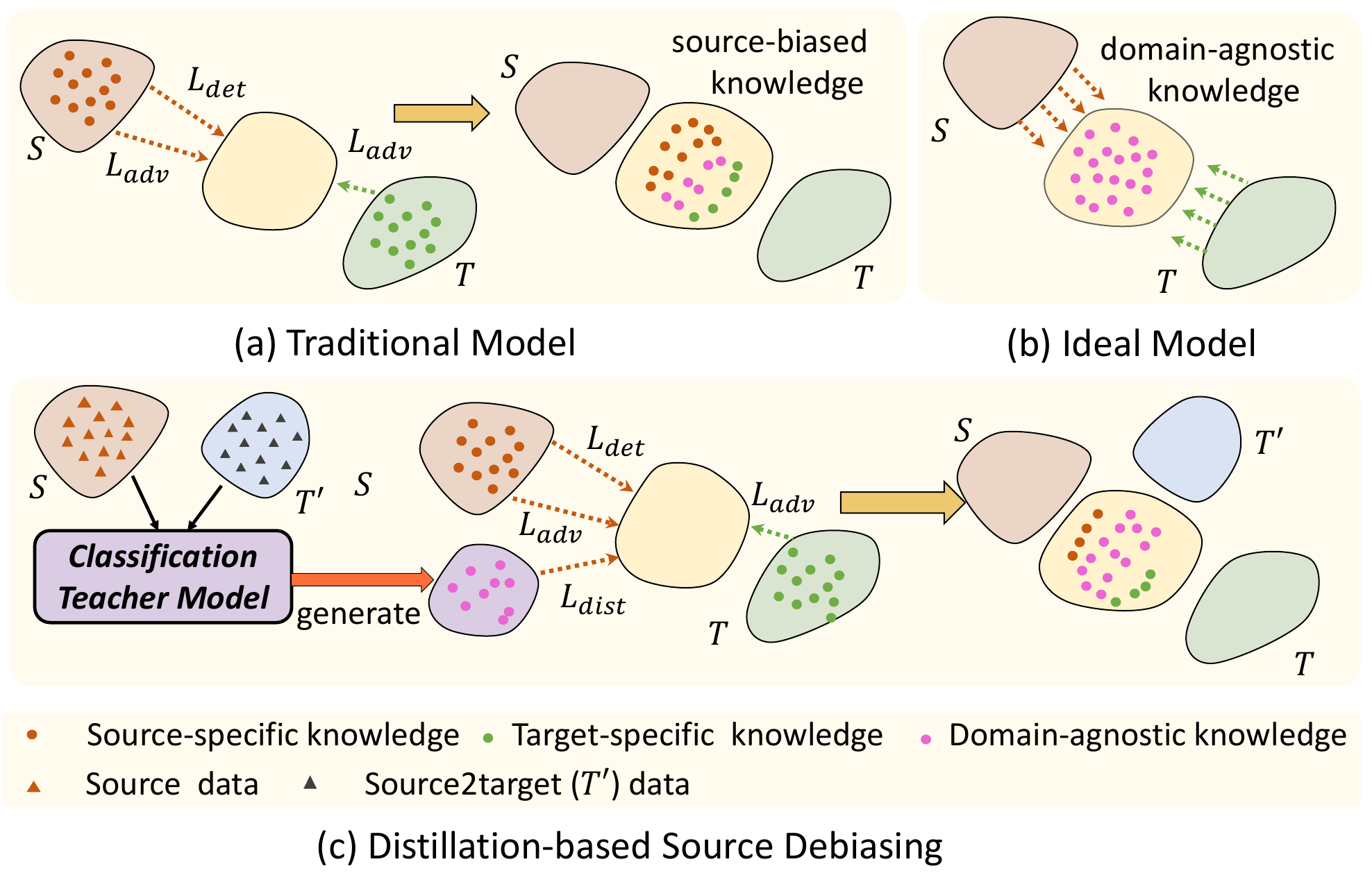} 
\caption{
In (a) traditional alignment approaches, the detector tends to learn more source-specific knowledge due to the supervised detection loss $L_{det}$, rather than domain-agnostic knowledge (b). In (c) our proposed DSD framework, by introducing a distillation loss to the source data, the model can acquire more domain-agnostic knowledge than (a). The red and green dotted arrows represent the impact of source or target-related losses on knowledge transfer, respectively.
}
\label{fig1}
\end{figure}

Extensive researches have been dedicated to address the challenge via Unsupervised Domain Adaption (UDA) methods, which aim to adapt to unlabeled target domain using the annotated source domains. One of the fashionable frameworks of UDA  
is to align the feature distributions between the source and target domains toward a cross-domain feature space. Early researches \cite{chen2018domain,saito2019strong,hsu2020every,jiang2021decoupled} aim to align image-level and instance-level features and achieve great margins over plain detectors. Recent works \cite{tian2021knowledge,zhang2021rpn,li2022sigma} devote to aligning class-conditional distribution across different domains, and have achieved fine-grained adaption in a category-wise manner. These approaches expect the model supervised by the labeled source domain to infer on the target domain effectively.

Despite of great success, there are still two challenges in existing alignment-based methods \cite{chen2018domain,zheng2020cross,li2022cross}. 
On one hand, as illustrated in Fig.~\ref{fig1}(a), the supervision of the model originates from two aspects: 1) supervised detection loss $L_{det}$ from source; 2) adversarial loss $L_{adv}$ from both source and target. Compared with $L_{adv}$, $L_{det}$ provides more explicit supervision signal and enforce the model to fit the source distribution. Thus, in this process, it is inevitable that detector will acquire more source-specific knowledge than target-specific knowledge. Simultaneously, the detector also gain domain-agnostic knowledge in the adversarial training process. As a result, the knowledge acquired by the detector is source-biased (\textit{i.e.,} more source-specific knowledge) rather than ideal (Fig.~\ref{fig1}(b)), hindering the model's generalization in the target domain.
These observations motivate us to design a new paradigm that enables the model to learn more domain-agnostic knowledge compared to traditional methods.

\begin{figure}[t]
\centering
\includegraphics[width=1.00\columnwidth]{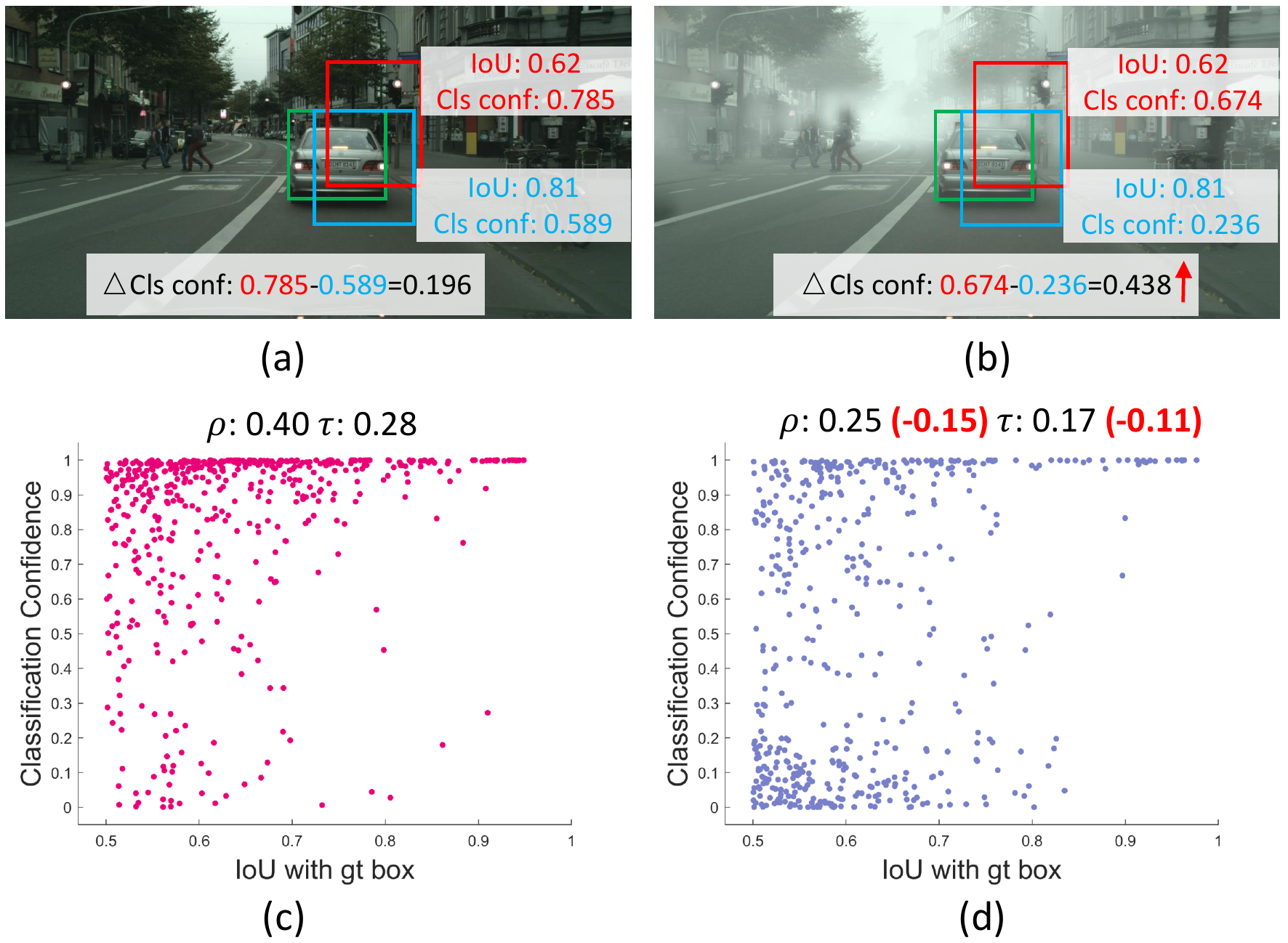} 
\caption{
Demonstrative cases of the \textbf{exacerbated inconsistency} between classification and localization based on alignment-based detector DA-Faster \cite{chen2018domain}. The upper row figures (a) and (b) show DA-Faster's detection results on \textit{Cityscapes} and \textit{FoggyCityscapes}, respectively. 
The lower row figures (c) and (d) display the correlation between localization ground-truth ($x$-axis, represented by the the IoU between the bounding box and its matched ground-truth) and classification scores ($y$-axis) for 500 randomly sampled DA-Faster's detected boxes in \textit{Cityscapes} and \textit{FoggyCityscapes}, respectively. The bounding boxes are filtered based on an IoU ($\ge 0.5$) with the corresponding ground-truth.}
\label{fig2}
\end{figure}

On the other hand, we observed that the alignment-based detectors (e.g. DA-Faster \cite{chen2018domain}) face \textbf{exacerbated inconsistency} between classification and localization. Firstly, as shown in Fig.~\ref{fig2}(a), inconsistency means the phenomenon \cite{jiang2018acquisition} that compared with the detected bounding box (blue), another detected bounding box (red) with higher classification scores could have lower IoU with ground truth boxes (green). \textbf{Exacerbated inconsistency} in this paper means that existing alignment-based detectors encounter more severe inconsistency issues on the target domain than that of source domain. 
From the perspective of detection visualization, the exacerbated inconsistency is reflected in that detection boxes (blue and red) in \textit{FoggyCityscapes} (Fig.~\ref{fig2}(b)) often exhibit larger differences in classification scores compared with ones located in the same position in \textit{Cityscapes} (Fig.~\ref{fig2}(a)). 
From the perspective of correlation metric, the exacerbated inconsistency means Spearman Rank Correlation Coefficient $\rho$ and Kendall Rank Correlation Coefficient $\tau$ in the \textit{FoggyCityscapes} (Fig.~\ref{fig2}(d)) are lower than ones in \textit{Cityscapes} (Fig.~\ref{fig2}(c)). $\rho$ and $\tau$ are the measures of rank correlation: the similarity of the ordings of the data when ranked by each of the quantities. 
As shown in Fig.~\ref{fig2}(c) (d), the $\rho$ and $\tau$ in the \textit{FoggyCityscapes} are relatively lower (-0.15 and -0.11) compared to \textit{Cityscapes}, indicating that the detector faces a more pronounced inconsistency issue in the target domain.
In general detection pipelines, the classification scores are commonly employed as the metric for ranking the detected boxes, which can result in the suppression of accurate bounding boxes by less accurate ones during NMS procedure. In the cross-domain detection, this issue is amplified due to more severe inconsistency.

In this paper, to overcome the aforementioned constraints, we present a DSD-DA method that contains a novel Distillation-based Source Debiasing (DSD) framework and a Domain-aware Consistency Enhancing (DCE) strategy. 
Firstly, considering that current alignment-based DAOD methods tends to acquire more source-specific knowledge, we propose to perform knowledge distillation to guide detector to acquire more domain-agnostic knowledge and thus restrain the potential source bias issue. Specifically, as shown in Fig.~\ref{fig1}(c), we first utilize CycleGAN \cite{zhu2017unpaired} to transform the source images into the target domain style, named source2target domain ${T'}$ (light blue). Then, we train a classification-teacher via a supervised and adversarial loss, which is able to learn domain-agnostic knowledge from $S$ and $T'$ mix-style data. Finally, these knowledge is utilized to guide the detector's training, improving the performance of detector on both domains.
Secondly, to mitigate exacerbated inconsistency on target data, we design a Target-Relevant Object Localization Network (TROLN), in which pixel and instance-level target affinity weights are proposed and incorporated into the loss function. TROLN is also trained on $S$ and $T'$ mix-style data, which is utilized to mine target relevant localization information. Then, in the testing stage, we conduct the DCE strategy, \textit{i.e.,} formulating the output of TROLN into a new localization representation. And we use this representation to adjust the classification scores of the detected boxes, enhancing the consistency and making sure that more accurate detected boxes are retained in NMS process.

In summary, our contributions are as follows:

\begin{itemize}
  \item We propose a novel DSD framework for DAOD, which utilizes an unbiased classification-teacher to guide the detector to learn more domain-agnostic feature representations. To the best of our knowledge, this is the first study to analyze and solve source bias issue in alignment-based methods.
  \item We reveal the \textbf{exacerbated inconsistency} issue between the classification and localization existing in traditional alignment-based method. We design TROLN and conduct DCE strategy to refine classification scores in the testing stage, which enhances the consistency between the classification and localization.
  \item Extensive experiments demonstrate that our method consistently outperforms the strong baseline by significant margins, highlighting its superiority compared to existing alignment-based methods.
\end{itemize}

\section{Related Work}

\textbf{Domain adaptation for object detection.} 
Several approaches have been proposed for DAOD, which can be categorized into alignment-based \cite{chen2018domain, saito2019strong, li2022sigma,li2022scan, xu2022h2fa} and self-training \cite{deng2021unbiased, ramamonjison2021simrod,deng2023harmonious} methods. 
However, regardless of various technological approaches, the source bias issue persists. Self-training methods alleviate the detector's bias towards the source domain by continuously improving the quality of pseudo-labels during training stage. The development of alignment-based methods often involves modeling features from coarse to fine. DA-Faster \cite{chen2018domain} implements feature alignment at both the image-level and instance-level. 
$H^2FA$ \cite{xu2022h2fa} enforces two image-level alignments for the backbone features, as well as two instance-level alignments for the RPN
and detection head.
SIGMA \cite{li2022sigma}  constructs the feature distributions of the source and target domains as graphs and reformulates the adaption with graph matching. However, regardless of the granularity of modeling, it is unable to change the asymmetry in the losses in the alignment stage and thus cannot effectively address the source bias problem. 
In this paper, we propose a novel distillation-based alignment (DSD) framework, where a distillation loss is constructed to guide the learning process of detector. To the best of our knowledge, this is the first method directly optimizing source bias in alignment-based approaches.

\textbf{Representation of localization quality.} The conflict between classification and localization tasks is a well-known problem \cite{jiang2018acquisition,tychsen2018improving,li2020generalized,zhang2021varifocalnet,pu2023rank,zhang2023decoupled, feng2021tood} in the object detection field. 
Existing works have focused on finding more accurate localization representation to guide the learning of the classification head, addressing the inconsistency issue.
IoU-Net \cite{jiang2018acquisition} introduces an extra head to predict IoU and use it to rank bounding boxes in NMS. Fitness NMS \cite{tychsen2018improving} and  IoU-aware RetinaNet \cite{wu2020iou} multiply the predicted IoU or IoU-based ranking scores by the classification score as the ranking basis. Instead of predicting the IoU-based score, FCOS \cite{tian2019fcos} predicts centerness scores to suppress the low-quality detections.
However, for exacerbated inconsistency in cross-domain scenarios, how to incorporate target-relevant information into localization representation designing has become a new challenge. 
Unfortunately, few methods explore and solve this challenge.
Therefore, we first propose a target-relevant OLN to mine target-related localization information from style-mixed data. Then we integrate these target relevant information into a novel localization representation to refine the classification scores, enhancing the consistency.

\begin{figure*}[t]
\centering
\includegraphics[width=1.00\textwidth]{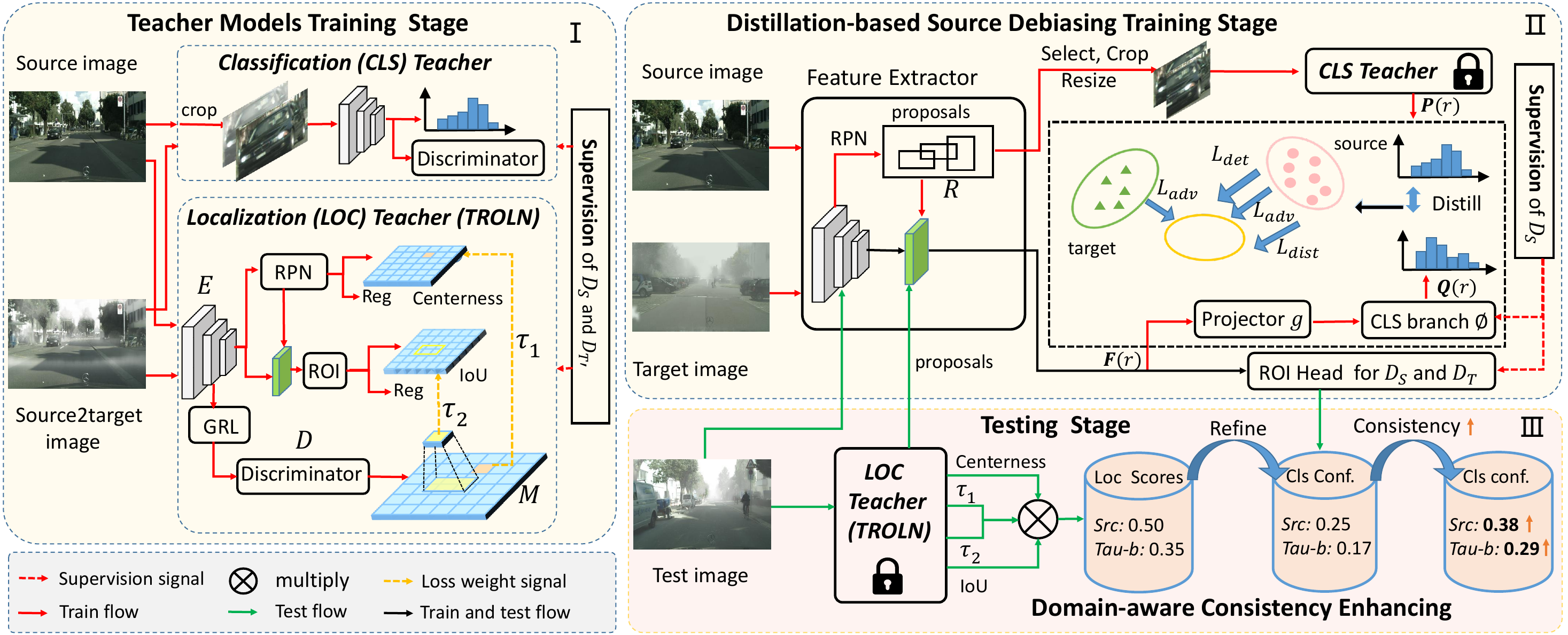} 
\caption{Overview of the proposed distillation-based source debiasing framework for DAOD.
Part \Rmnum{1} shows the teacher models training stage, which includes a mix-style classifier and a Target-Relevant object localization network (TROLN) training. Part \Rmnum{2} demonstrates distillation-based source debiasing (DSD) training, in which the cross-domain detector is trained. In Part \Rmnum{3}, the Domain-aware Consistency Enhancement (DCE) strategy is introduced to refine the detector's classification scores in the testing phase, enhancing the consistency between classification and localization.}
\label{fig3}
\end{figure*}

\section{Method}

\subsection{Problem Formulation}
In the cross-domain object detection, we have a labeled source domain $\mathcal{D}_S = \{ (x_i^s, y_i^s) \}_{i=1}^{N_s}$, where $x_i^s$ and $y_i^s=(b_i^s, c_i^s)$ denote the $i_{th}$ image and its corresponding labels, \textit{i.e.}, the coordinates of the bounding box $b$ and its associated category $c$, respectively. In addition, we have access to an unlabeled target domain $\mathcal{D}_T = \{ x_i^t \}_{i=1}^{N_t}$. 
In this work, we employ CycleGAN \cite{zhu2017unpaired} to convert the source images into the target domain style, creating a new domain named source2target domain $\mathcal{D}_{T'} = \{ (x_i^{t'}, y_i^{t'}) \}_{i=1}^{N_s}$, which shares labels with the source domain data. 
We assume that the source and target samples come from different distributions (\textit{i.e.}, $\mathcal{D}_S \neq \mathcal{D}_T$) but the categories are exactly the same. The objective is to enhance the performance of the detector in $\mathcal{D}_T$ using the knowledge in $\mathcal{D}_S$. 

\subsection{Framework Overview}

As shown in Fig.~\ref{fig3}, our method involves two training stages, a teacher models training stage and a detector training stage. In the teacher models training stage (Sec~\ref{pretraining}), we train a classification and a localization models as teachers using the labeled data $\mathcal{D}_S$ and $\mathcal{D}_{T'}$. In the second training stages (Sec~\ref{distillation}), the features of the positive proposals are expected to derive domain-agnostic representation from the classification-teacher model. During the testing stage (Sec~\ref{localizaion}), we design the localization scores based on the output of TROLN and use it to refine the classification scores, thereby alleviating the inconsistent issue. 

\subsection{Detection Baseline}
We use DA-Faster \cite{chen2018domain} as our base detector. DA-Faster is a two-stage cross-domain detector that consists of two major components: a standard Faster-RCNN detector and a domain adaption component that includes image-level and instance-level domain discriminators $D$. When the training process gradually converges, the detector tends to extract domain-invariant feature representations. 
Formally, the image-level adaption loss can be written as:
\begin{equation}
   \begin{split}
      \mathcal{L}_{img} = \min \limits_{\theta_{G}} \max \limits_{\theta_{D}} \ & \mathbb{E}_{x_s \sim \mathcal{D}_S} \log D_{img}(G(x_s)) \\ 
       + \ & \mathbb{E}_{x_t \sim \mathcal{D}_T} \log (1 - D_{img}(G(x_t)))
   \end{split}
\end{equation}
where $\theta_{G}$ and $\theta_{D}$ are the parameters of backbone $G$ and $D_{img}$, $x_s$, $x_t$ represent images from ${\mathcal{D}_S}$ and ${\mathcal{D}_T}$, respectively. Similarly, the instance-level adaption loss is defined as:
\begin{equation}
   \begin{split}
      \mathcal{L}_{ins} = \min \limits_{\theta_{Det}} \max \limits_{\theta_{D}} \ & \mathbb{E}_{f_s \sim {ROI}_S} \log D_{ins}(f_s) \\ 
       + \ & \mathbb{E}_{f_t \sim {ROI}_T} \log (1 - D_{ins}(f_t))
   \end{split}
\end{equation}
where $\theta_{Det}$ and $\theta_{D}$ are the parameters of detector and $D_{ins}$, $f_s$, $f_t$ represent the ROI features of $x_s$ and $x_t$, respectively.
The training loss of DA-Faster is a summation of each individual part, which can be written as:
\begin{equation}
    \mathcal{L}_{DA} = {L}_{det} + \lambda ({L}_{img} + {L}_{ins}) 
\end{equation}
where $\mathcal{L}_{det}$ is the loss of Faster-RCNN and $\lambda$ is a trade-off parameter to balance the Faster-RCNN loss and domain adaption loss.

\subsection{Teacher Models Training}
\label{pretraining}

\textbf{Classification Teacher}.
We first construct an instance-level image dataset $\mathcal{D}$ as image corpus by extracting all class objects from the detection dataset $\mathcal{D}_S$ and $\mathcal{D}_{T'}$ according to their ground-truth bounding boxes and labels. 
Formally, for an input image, we first perform typical data augmentations (random cropping, color distortion, \textit{etc.}). Then we feed the augmented image into a ResNet \cite{he2016deep} classification network for supervised learning. Simultaneously, we adopt domain discriminator to align the feature distribution of $S$ and $T'$.

The classifier's ability to acquire domain-agnostic knowledge can be attributed to three aspects:
1) Strong supervision signal. Since $\mathcal{D}$ contains labeled images with two different styles, classifier is enforced to fit the data distribution of different domain via the supervised loss during the training. In this process, classifier tends to acquire domain-agnostic knowledge.
2) Adversarial learning. Domain discriminator with Gradient Reverse Layer (GRL) \cite{ganin2015unsupervised} layer is powerful tool that can align feature distributions between two domains. It is beneficial for feature extractor to produce domain-invariant features that cannot be discriminated by the discriminator.
3) Enriched data. Compared with sole source data, the mixed data doubles the scale and greatly enriches the training data with various data augmentations. It facilitates the classifier to learn domain-invariant representations.
Our classification-teacher model is optimized in a completely independent way from the object detection. And its domain-agnostic knowledge can be transferred to object detection to suppress potential source bias.


\textbf{TROLN Teacher}.
To solve the challenge of exacerbated inconsistency on the target data, we attempt to use localization indicators (IoU, centerness) from OLN \cite{kim2022learning} to calibrate the classification scores. The original OLN estimates the objectness of each region by centerness-head and IoU-head. The comprehensive loss function of OLN can be written as:
\begin{equation}
\label{oln}
\begin{gathered}
    \mathcal{L}_{OLN} = \mathcal{L}_{RPN}^{Cent} + \mathcal{L}_{RPN}^{reg} + \mathcal{L}_{RCNN}^{IoU} + \mathcal{L}_{RCNN}^{reg} \\
    \mathcal{L}_{RPN}^{Cent} = \frac{1}{N_{pix}} \sum_{x=1}^{W} \sum_{y=1}^{H} \mathbb{1}_{for}^{pix} {L}_{1}(c_{x,y}, \hat{c}_{x,y}) \\
    \mathcal{L}_{RCNN}^{IoU} = \frac{1}{N_{pos}} \sum_{r=1}^{N_{pos}} \mathbb{1}_{for}^{pro} {L}_{1}(b_r, \hat{b}_r)
\end{gathered}
\end{equation}
where $\mathbb{1}_{for}^{pix}$ and $\mathbb{1}_{for}^{pro}$ denote the positive pixels and positive proposals set. $c_{x,y}$, $b_r$, $\hat{c}_{x,y}$, $\hat{b}_r$ are the predicted centerness, predicted IoU, groundtruth centerness and groundtruth IoU, respectively.  

However, when using directly $S$ and $T'$ mixed data to train original OLN, the lack of guidance from the target domain results in less target-relevant images being given the same importance as more relevant ones, leading to a deterioration in knowledge learning and mining from the target domain. 
To effectively mine target relevant localization information in the training, Target-Relevant Object Localization Network (TROLN) has been developed to ensure that target-relevant information are encoded at the pixel and instance level. Specifically, a pixel-level domain discriminator $D$ is placed after the feature encoder $E$ (shown in Fig.~\ref{fig3} \Rmnum{1}) in the TROLN.  
The probability of each pixel belonging to the target domain is defined as $D(E(X)) \in \mathbb{R}^{H \times W \times 1}$ and $1 - D(E(X)) \in \mathbb{R}^{H \times W \times 1}$ represents the probability of it belonging to the source domain. The domain discriminator $D$ is updated using binary cross-entropy loss based on the domain label $d$ for each input image, where images from the source domain are labeled as $d=0$ and images from target domain are labeled as $d=1$. The discriminator loss  $\mathcal{L}_{dis}$ can be expressed as:

\begin{equation}
\begin{split}
\mathcal{L}_{dis} = - d \log D(E(X)) 
- (1-d) \log (1-D(E(X))) 
\end{split}
\end{equation} 

The large value within $D(E(X))$ indicates that the distribution of current pixel and target pixels are more similar. Based on the important cues, we denote $D(E(X))$ as target affinity score map $M$ and adopt dynamic target-related weight to adjust the ${L}_{RPN}^{Cent}$ and ${L}_{RCNN}^{IoU}$. 
The pixel-level and instance-level target affinity weight $\tau_{1}$, $\tau_{2}$ are defined as the following:
\begin{equation}
\label{oln-t}
\begin{gathered}
    \tau_{1} = M(x, y) \\
    \tau_{2} = \text{Average}(\text{ROI}\text{Align}(M, p))
\end{gathered}
\end{equation}
where $(x,y)$ represents the pixel coordinates of $M$ and $p$ denotes one proposal from RPN.

Subsequently, we can reweight the importance of loss items from pixel and instance level as illustrated in Fig.~\ref{fig3} \Rmnum{1}, and apply it to train a localization-teacher (TROLN) by reformulating the loss function in Eq.~\ref{oln} as the following:
\begin{equation}
\label{myoln}
\begin{gathered}
    \mathcal{L}_{TROLN} = \mathcal{L}_{RPN}^{Cent} + \mathcal{L}_{RPN}^{reg} + \mathcal{L}_{RCNN}^{IoU} + \mathcal{L}_{RCNN}^{reg} + \mathcal{L}_{dis} \\
    \mathcal{L}_{RPN}^{Cent} = \frac{1}{N_{pix}} \sum_{x=1}^{W} \sum_{y=1}^{H} \mathbb{1}_{for}^{pix} (\tau_{1}+1){L}_{1}(c_{x,y}, \hat{c}_{x,y}) \\
    \mathcal{L}_{RCNN}^{IoU} = \frac{1}{N_{pos}} \sum_{r=1}^{N_{pos}} \mathbb{1}_{for}^{pro} (\tau_{2}+1){L}_{1}(b_r, \hat{b}_r)
\end{gathered}
\end{equation}

Based on Eq.~\ref{myoln}, TROLN is explicitly enforced to learn from target-relevant samples, and thus prevents the interference from the information irrelevant to the target.

\begin{figure}[t]
\centering
\includegraphics[width=0.45\textwidth]{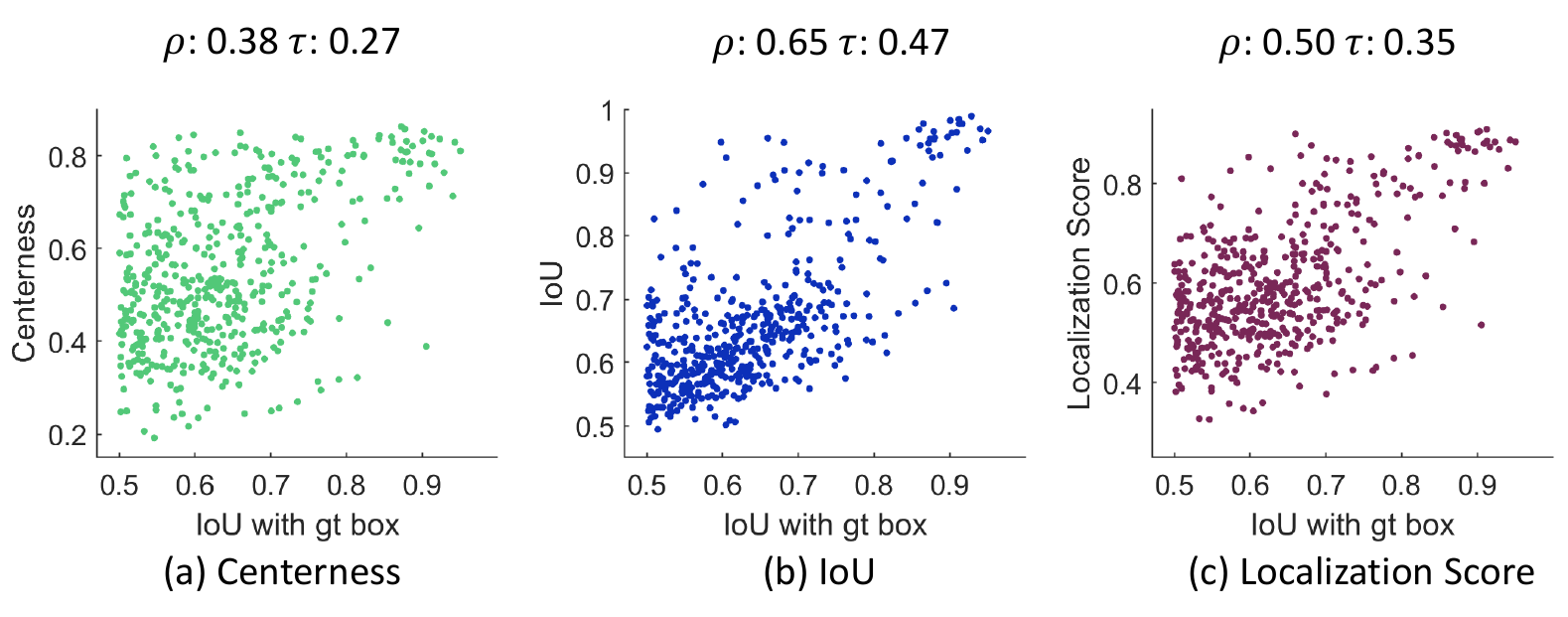} 
\caption{The correlation between localization ground-truth and centerness/IoU/localization scores of the bounding boxes on the target test dataset. The bounding boxes are filtered based on an IoU ($\ge 0.5$) with the corresponding ground-truth.}
\label{fig4}
\end{figure}

\subsection{Distillation-based Source Debiasing}
\label{distillation}

After Teacher Models Training stage, we start the training of cross-domain detector.
Here, we take DA-Faster \cite{chen2018domain} as the base detector to describe the DSD framework.
As shown in Fig.~\ref{fig3} \Rmnum{2}, in the DSD training stage, given source proposals $R$ generated by RPN, we initially select $R$ by assessing IoU with ground truth higher than threshold $T$. We crop these proposals from the source image and resize them to a fixed size using bilinear interpolation, then feed them into our classification-teacher model to obtain classification logit $\mathbf{P}(r)\in \mathbb{R}^{K\times 1}$. Here, $K$ represents the number of classes in the object detection task, $r$ represents one of filtered proposals.
Meanwhile, we obtain the ROI features $\mathbf{F}(r)$ for $r$ from the RoI Align layer. Note that $\mathbf{F}(r)$ and $\mathbf{P}(r)$ are learned in the different feature space, thus we first project $\mathbf{F}(r)$ into the same feature space as $\mathbf{P}(r)$ and then obtain the classification logit of the projected feature:
\begin{equation}
\begin{split}
   \mathbf{Q}(r) = \phi(g(\mathbf{F}(r))
  \label{eq:student}
  \end{split}
\end{equation}
Here, $g(\cdot)$ denotes the project function for features $\mathbf{F}(r)$, which is implemented with a $1\times1$ convolutional layer, while $\phi(\cdot)$ is the classification branch in the detection head.
Then we minimize the L1-norm between these two logit representations to guide the learning process of the detector:
\begin{equation}
  L_\text{dist} = \frac{1}{{R}_{f}K}\sum^{{R}_{f}}_{r=1}\sum^{K}_{k=1}\left\| \mathbf{P}_{k}(r)-\mathbf{Q}_{k}(r) \right\|_1
  \label{eq:distillation}
\end{equation}
where ${R}_{f}$ is the number of filtered proposals. 
The obtained logit representation $\mathbf{Q}(r)$ can also be utilized for classification of the region proposal $r$. Thus, we conduct an auxiliary classification task on the logit $\mathbf{Q}(r)$.
\begin{equation}
\begin{split}
  & p^{\prime} = \mathcal{F}_{\text{softmax}}(\mathbf{Q}(r)) \\
  & \mathcal{L}_\text{cls-aux} = \text{CE}(y, p^{\prime}),
  \end{split}
  \label{eq:score}
\end{equation}
where $p^{\prime}$ is the predicted scores based on $\mathbf{Q}(r)$, $y$ is the groundtruth label for the region proposal $r$. Note that the whole distillation process is only conducted on the source images. 

Consequently, the object detector is trained under the supervision of the three losses jointly:
\begin{equation}
   \mathcal{L}_\text{obj} = \mathcal{L}_\text{DA} + \mathcal{L}_\text{dist} + \mathcal{L}_\text{cls-aux},
  \label{eq:loss_student}
\end{equation}
where $\mathcal{L}_\text{DA}$ denotes the loss of DA-Faster.

\subsection{Domain-aware Consistency Enhancing}
\label{localizaion}

To address more severe inconsistency between classification and localization, 
we design a novel localization representation to refine classification scores, enhancing the consistency in the testing stage. Since the objective of DAOD is to enhance the performance of the detector in target domain and TROLN has already explored the target-relevant localization information, we first investigate the relation between localization ground-truths and two localization indicators (IoU and centerness) derived from TROLN.

In the TROLN framework, each detected bounding box originates from two refinements of the corresponding anchor, \textit{i.e.,} anchor ${\rightarrow}$ proposal ${\rightarrow}$ detected box. Here, we assume that these three share the same centerness and IoU.
Similar to Fig.~\ref{fig2}, we evaluate the trained TROLN on the \textit{FoggyCityscapes} test datasets and visualize the correlation of two indicators ($y$-axis, centerness and IoU) and localization ground-truth ($x$-axis), as shown in Fig.~\ref{fig4}(a) and (b).
It is evident that compared to the classification score in Fig.~\ref{fig2}(b), both IoU and centerness exhibit a higher consistency with the localization ground-truths. 
This indicates that these two indicators have the potential to serve as localization representation for refining the classification scores.

However, on one hand, using the highest consistency indicator (IoU) to weight the classification scores could cause an extra class confusion issue. For example, if two detected boxes simultaneously match the same ground truth box with the category ``cat", where the detected box \textit{A} predicts a ``cat" confidence of 0.6 and an IoU of 0.5, and detected box \textit{B} predicts a ``tiger" confidence of 0.5 and an IoU of 0.7. In this case, employing IoU would yield a ``cat" confidence of 0.3 ($0.6\times0.5$) for detected box \textit{A} and a "tiger" confidence of 0.35 ($0.5\times0.7$) for detected box \textit{B}, potentially resulting in misclassification.
On the other hand, refining classification score based on centerness may not sufficiently improve the consistency. 
In addition, although centerness and IoU have already integrate target-relevant information, they lack the capability to adaptively adjust based on current features. Therefore, we incorporate pixel and instance-level target affinity weights $\tau_{1}$, $\tau_{2}$ into the localization representation and strike a balance between centerness and IoU. Here, we propose a novel localization score $s$ as follow:
\begin{equation}
   s = \sqrt{4\times{c}\times{b}\times\tau_{1}\times\tau_{2}},
\label{eq:s}
\end{equation}

where $c$ and $b$ denote centerness and IoU, respectively. Due to the effect of the GRL in TROLN, $\tau_{1}$ and $\tau_{2}$ are close to ``0.5'' with the training process gradually converge, the ``4'' in Eq.~\ref{eq:s} is a compensation factor.
From Fig.~\ref{fig4}(c), we can observe that the consistency of $s$ ($\rho$ and $\tau$) ranges between these of $c$ and $b$.
The rationale behind Eq.~\ref{eq:s} is to improve the localization scores of target-style detection boxes (with larger $\tau_{1}$, $\tau_{2}$) and suppress source-style ones (with smaller $\tau_{1}$, $\tau_{2}$) during the testing stage. This operation is to adaptively retain more target-style detected boxes.

\begin{table*}
    \caption{Results from \textbf{\textit{Cityscapes$\rightarrow$FoggyCityscapes}} based on different base detectors with various backbones.}
        \centering
         \resizebox{1.0\textwidth}{!}{%
        \begin{tabular}{>{\raggedright}p{6.7cm}|p{2.0cm}<{\centering}|p{1.2cm}<{\centering}p{1.2cm}<{\centering}p{1.2cm}<{\centering}p{1.2cm}<{\centering}p{1.2cm}<{\centering}p{1.2cm}<{\centering}p{1.2cm}<{\centering}p{1.2cm}<{\centering}p{1.2cm}<{\centering}}
        \hline
        Method & Backbone & person & rider & car & truck & bus & train & mcycle & bcycle & mAP \\ \hline \hline
        SWDA \cite{saito2019strong}  & \multirow{14}{*}{VGG16} & 36.2 & 35.3 & 43.5 & 30.0 & 29.9 & 42.3 & 32.6 & 24.5 & 34.3 \\
        TIA \cite{zhao2022task} &  & 34.8 & 46.3 & 49.7 & 31.1 & 52.1 & 48.6 & 37.7 & 38.1 & 42.3 \\
        TDD \cite{he2022cross} &  & 39.6 & 47.5 & 55.7 & 33.8 & 47.6 & 42.1 & 37.0 & 41.4 & 43.1 \\
        MGA \cite{zhou2022multi} &  & 45.7 & 47.5 & 60.6 & 31.0 & 52.9 & 44.5 & 29.0 & 38.0 & 43.6 \\
        PT \cite{chen2022learning} &  & 40.2 & 48.8 & 59.7 & 30.7 & 51.8 & 30.6 & 35.4 & 44.5 & 42.7 \\
        SCAN \cite{li2022scan} &  & 41.7 & 43.9 & 57.3 & 28.7 & 48.6 & 48.7 & 31.0 & 37.3 & 42.1 \\
        SIGMA++ \cite{li2023sigma++} &  & 46.4 & 45.1 & 61.0 & 32.1 & 52.2 & 44.6 & 34.8 & 39.9 & 44.5 \\
        HT \cite{deng2023harmonious} &  & 52.1 & 55.8 & 67.5 & 32.7 & 55.9 & 49.1 & 40.1 & 50.3 & 50.4 \\
        OADA \cite{yoo2022unsupervised} &  & 47.8 & 46.5 & 62.9 & 32.1 & 48.5 & 50.9 & 34.3 & 39.8 & 45.4 \\
        SIGMA \cite{li2022sigma} &  & 43.9 & 52.7 & 56.8 & 26.2 & 46.2 & 12.4 & 34.8 & 43.0 & 43.5 \\
        DA-Faster\cite{chen2018domain} &  & 43.9 & 52.9 & 56.8 & 26.2 & 46.2 & 12.4 & 34.9 &43.0 & 39.6 \\
        \cellcolor{lightgray!45}DA-Faster\cite{chen2018domain} + DSD-DA & &  \cellcolor{lightgray!45}46.5 & \cellcolor{lightgray!45}54.1 & \cellcolor{lightgray!45}61.9 & \cellcolor{lightgray!45}28.3 & \cellcolor{lightgray!45}49.5 & \cellcolor{lightgray!45}26.7 & \cellcolor{lightgray!45}40.0 & \cellcolor{lightgray!45}46.3 & \cellcolor{lightgray!45}\textbf{44.2}  \\ 
       
        AT \cite{li2022cross} &  &  45.3 & 55.7 & 63.6 & 36.8 & 64.9 & 34.9 & 42.1 & 51.3 & 49.3  \\ 
          \cellcolor{lightgray!45}AT \cite{li2022cross} + DSD-DA  & &  \cellcolor{lightgray!45}49.1 & \cellcolor{lightgray!45}59.3 & \cellcolor{lightgray!45}66.2 & \cellcolor{lightgray!45}35.8 &  \cellcolor{lightgray!45}60.0 &  \cellcolor{lightgray!45}47.1 & \cellcolor{lightgray!45}45.2 & \cellcolor{lightgray!45}54.9 & \cellcolor{lightgray!45}\textbf{52.2}  \\  
        CMT \cite{cao2023contrastive} & \multirow{2}{*} &  45.9 & 55.7 & 63.7 & 39.6 &66.0 & 38.8 & 41.4 & 51.2 & 50.3  \\ 
         \cellcolor{lightgray!45}CMT \cite{cao2023contrastive} + DSD-DA  & &   \cellcolor{lightgray!45}49.0 &  \cellcolor{lightgray!45}59.6 &  \cellcolor{lightgray!45}65.3 &  \cellcolor{lightgray!45}35.7 &   \cellcolor{lightgray!45}61.0 &   \cellcolor{lightgray!45}46.5 &  \cellcolor{lightgray!45}43.9 &  \cellcolor{lightgray!45}57.3 &  \cellcolor{lightgray!45}\textbf{52.3}  \\  
        \hline \hline
        GPA \cite{xu2020cross} & \multirow{6}{*}{ResNet50} & 32.9 & 46.7 & 54.1 & 24.7 & 45.7 & 41.1 & 32.4 & 38.7 & 39.5 \\
        CRDA \cite{xu2020exploring} &  & 39.9 & 38.1 &57.3 & 28.7 & 50.7 & 37.2 & 30.2 & 34.2 & 39.5 \\
        DIDN \cite{lin2021domain} &  & 38.3 & 44.4 & 51.8 & 28.7 & 53.3 & 34.7 & 32.4 & 40.4 & 40.5 \\
        DSS \cite{wang2021domain} &  &42.9 & 51.2 & 53.6 &  33.6 & 49.2 & 18.9 & 36.2 & 41.8 & 40.9 \\
        DA-Faster\cite{chen2018domain} &  & 36.4 & 47.2 & 53.7 & 29.3 & 48.8 & 34.4 & 33.8 & 38.5 & 40.2 \\
        \cellcolor{lightgray!45}DA-Faster\cite{chen2018domain} + DSD-DA & &  \cellcolor{lightgray!45}43.7 & \cellcolor{lightgray!45}49.1 & \cellcolor{lightgray!45} 60.7 &\cellcolor{lightgray!45}30.8 & \cellcolor{lightgray!45}55.7 & \cellcolor{lightgray!45}43.4 & \cellcolor{lightgray!45}33.7 & \cellcolor{lightgray!45}44.6 & \cellcolor{lightgray!45}\textbf{45.2} \\
        \hline \hline
        
        CADA \cite{hsu2020every} & \multirow{4}{*}{ResNet101} & 41.5 & 43.6 & 57.1 & 29.4 & 44.9 & 39.7 & 29.0 & 36.1 & 40.2 \\
        D-adapt \cite{jiang2021decoupled} &  & 42.8 & 48.4 & 56.8 & 31.5 & 42.8 & 37.4 & 35.2 & 42.4 & 42.2 \\
        DA-Faster\cite{chen2018domain} &  & 37.2 & 45.1 & 54.5 & 30.9 & 48.9 & 43.3 & 29.3 & 39.5 & 41.1 \\
        \cellcolor{lightgray!45}DA-Faster\cite{chen2018domain} + DSD-DA & &\cellcolor{lightgray!45}43.9 &  \cellcolor{lightgray!45}50.7 & \cellcolor{lightgray!45}61.6 & \cellcolor{lightgray!45}31.8 & \cellcolor{lightgray!45}52.2 & \cellcolor{lightgray!45}47.1 & \cellcolor{lightgray!45}32.1 &\cellcolor{lightgray!45}46.1 & \cellcolor{lightgray!45}\textbf{45.7} \\
       \hline
        \end{tabular}
        }
        
        \label{table:foggy}
        
\end{table*}

\begin{table}
 \vspace{-0.4cm}
\caption{Results on \textbf{\textit{Kitti$\rightarrow$Cityscapes}} with VGG-16. SO represents the source only results and GAIN indicates the adaption gains compared with the source only model.}
    \centering
   
    \resizebox{0.48\textwidth}{!}{%
    \begin{tabular}{p{5.7cm}|cl}
    \hline
      Method                  &  Car                      & SO/GAIN       \\ 
    \hline \hline
    CADA \cite{hsu2020every} &  43.2   & 34.4/ 8.8\\
    MEGA \cite{vs2021mega}& 43.0   & 30.2/ 12.8\\
    SSAL \cite{munir2021ssal} &  45.6   & 34.9/ 10.7\\
    KTNet \cite{tian2021knowledge} &  45.6   & 34.4/ 11.2\\
    SIGMA \cite{li2022sigma} &  45.8   & 34.4/ 11.4\\
    DA-Faster \cite{chen2018domain} &  43.4   & 34.5/ 8.9\\
    \rowcolor{lightgray!45}DA-Faster \cite{chen2018domain}+ DSD-DA              & \textbf{46.9} & 34.5/ 12.4\\
    
    AT \cite{li2022cross} &  47.7   & 34.5/ 13.2\\
    \rowcolor{lightgray!45}AT \cite{li2022cross} + DSD-DA              & \textbf{49.3} & 34.5/ 14.8\\
    \hline
    \end{tabular}
    }
    
    \label{table:kitti}
    \vspace{-0.5cm}
\end{table}

\begin{table}[h]
\vspace{-0.4cm}
 \caption{Results from \textbf{\textit{SIM10k$\rightarrow$Cityscapes}}.}
 \centering
 \resizebox{0.47\textwidth}{!}{%
\begin{tabular}{l|cl}
\hline
Method       &  Car  & SO/GAIN \\ \hline\hline
SWDA \cite{inoue2018cross}       &      40.1                         &   34.3/ 5.8    \\ 
MAF  \cite{he2019multi}        &      41.1                         &   34.3/ 6.8   \\ 
HTCN \cite{chen2020harmonizing}        &    42.5                           &  34.4/ 8.1     \\
CFFA \cite{zheng2020cross}        &      43.8                         &  34.3/ 9.5     \\
ATF  \cite{he2020domain}        &      42.8                         &   34.3/ 8.5    \\
MeGA-CDA \cite{vs2021mega}    &     44.8                          &  34.3/ 10.5     \\
UMT \cite{deng2021unbiased} &       43.1                        &   34.3/ 8.8    \\  
DA-Faster \cite{chen2018domain} &  43.6   & 34.6/ 9.0\\
\rowcolor{lightgray!45}DA-Faster \cite{chen2018domain} + DSD-DA       &          \textbf{47.8}                    &   34.6/ 13.2     \\
AT \cite{li2022cross}          &   51.4      &   34.6/ 16.8       \\
\rowcolor{lightgray!45}AT \cite{li2022cross} + DSD-DA    &    \textbf{52.5}                           &   34.6/ 17.9       \\ \hline
\end{tabular}
}
\label{table:sim}
\vspace{-0.5cm}
\end{table}

Then we use $s$ to refine classification scores, termed as the Domain-aware Consistency Enhancing strategy.
Concretely, in the testing stage, given an image $I$, we feed it to the TROLN to obtain a proposals set $\mathcal{R} = \{ (box_i, s_i) \}_{i=1}^{N_p}$, where $box_i$ and $s_i$ represent the spatial coordinates and the localization score of the $i$-th proposal, $N_p$  denotes the total number of proposals. 
Simultaneously, we feed $I$ into the trained detector, replacing the detector's proposals with $\mathcal{R}$. As a result, we obtain the ROI head output $\mathcal{T} = \{ (reg_i, cls_i) \}_{i=1}^{N_p}$, where $reg_i$ and $cls_i$ denote the regression results and classification scores respectively. 
After experimenting with various forms such as squaring and other transformations, we ultimately adopt Eq. \ref{eq:adjust} to refine $cls_i$, and obtain the adjusted classification score $cls_{i}^{'}$:
\begin{equation}
  {cls_{i}^{'}} = \mathcal{F}_{\text{softmax}}\sqrt[4]{{cls_{i}}\times{s_{i}}}
  \label{eq:adjust}
\end{equation}
The refined output $\mathcal{T^{'}} = \{ (reg_i, cls_{i}^{'}) \}_{i=1}^{N_p}$ are used to participate NMS and evaluate the performance by following DA-Faster \cite{chen2018domain} protocol.

\section{Experiments}

\subsection{Datasets and Implementation Details}
We conduct our experiments on four datasets, including (1) 
\textbf{\textit{Cityscapes}}~\cite{cordts2016cityscapes} contains authentic urban street scenes captured under normal weather conditions, encompassing 2,975 training images and 500 validation images with pixel-level annotations. (2) \textbf{\textit{FoggyCityscapes}}~\cite{sakaridis2018semantic} is a derivative dataset that simulates dense foggy conditions based on Cityscapes, maintaining the same train/validation split and annotations. (3) \textbf{\textit{KITTI}}~\cite{geiger2012we} is one popular dataset for autonomous driving including 7,481 labeled images for training. (4) \textbf{\textit{SIM10k}}~\cite{johnson2016driving} is a synthetic dataset containing 10,000 images rendered from the video game Grand Theft Auto V (GTA5).

We report $AP_{50}$ of each class for object detection following \cite{chen2018domain} for all experimental setting as follows: (1) \textbf{\textit{Cityscapes$\rightarrow$FoggyCityscapes}}. It aims to perform adaptation across different weather conditions. 
(2) \textbf{\textit{Kitti$\rightarrow$Cityscapes}}. It is cross camera adaption, where the source and target domain data are captured with different camera setups. 
(3) \textbf{\textit{SIM10k$\rightarrow$Cityscapes}}. To adapt the synthetic scenes to the real one, we utilize the entire \textbf{\textit{SIM10k}} dataset as the source domain and the training set of \textbf{\textit{Cityscapes}} as the target domain. 
Following \cite{li2023sigma++}, we only report the performance on car for the last two scenarios.


\begin{table}[]
 \vspace{-0.1cm}
\caption{Detection performance on the source and target domain using different backbones. ${AP}_{s}$/${AP}_{t}$: Detection performance (AP) on the source/target.}
\centering
\begin{tabular}{p{2.0cm}|p{1.2cm}<{\centering}|p{0.75cm}p{0.75cm}<{\centering}}
\hline
Method      & \multicolumn{1}{c|}{Backbone} & ${AP}_{s}$$\uparrow$ & ${AP}_{t}$$\uparrow$   \\ \hline \hline
\textit{Source Only} & \multirow{3}{*}{VGG16}          & 49.02       & 20.18           \\  
\textit{Baseline}    &                               & 48.91       & 39.56            \\  
\textit{Baseline}+\textit{DSD}      &                               & \textbf{50.09}      &  \textbf{42.00}   \\ \hline
\textit{Source Only} & \multirow{3}{*}{ResNet50}     & 50.12       & 23.92           \\ 
\textit{Baseline}    &                               & 50.21       & 40.90           \\ 
\textit{Baseline}+\textit{DSD}      &                               & \textbf{51.48}       & \textbf{43.05}     \\ \hline
\end{tabular}
\label{table:bias}
\vspace{-0.5cm}
\end{table}

We evaluate the proposed DSD-DA method (DSD + DCE) on DA-Faster \cite{chen2018domain}, AT \cite{li2022cross} and CMT \cite{cao2023contrastive}. 
In the DSD training stage, we resize all the cropped images to 224$\times$224, and set IoU threshold $T=0.8$. DA-Faster was trained with SGD optimizer with a 0.001 learning rate, 2 batch size, momentum of 0.9, and weight decay of 0.0005 for 70k iterations on 1 Nvidia GPU 2080Ti. 

\subsection{Main Results}

\textbf{\textit{Cityscapes$\rightarrow$FoggyCityscapes}}. We present the comparison with VGG16, ResNet50 and ResNet101 backbones in Table~\ref{table:foggy}. When base detector is DA-Faster, our method achieves 44.2\%, 45.2\%, and 45.7\% mAP, respectively, improving mAP by 4.6\%, 5.0\% and 4.6\% compared to \cite{chen2018domain}. Simultaneously, our method has achieved consistent improvements on the state of the art such as AT \cite{li2022cross} and CMT \cite{cao2023contrastive}. This fully demonstrates the effectiveness of our approach and its compatibility with the different backbone networks.

\textbf{\textit{Kitti$\rightarrow$Cityscapes}}. In Table~\ref{table:kitti}, we illustrate the performance comparison on the cross-camera task. The proposed method reaches an $AP_{50}$ of 46.9\% and 49.3\% with a gain of +12.4\% and +14.8\% over the source only model with different base detector, respectively.

\textbf{\textit{SIM10k$\rightarrow$Cityscapes}}.
Table~\ref{table:sim} shows that our method consistently improves performance across different base detectors. This further illustrates that the proposed approach has strong generalization capabilities, effectively adapting from synthetic to real setting.



\subsection{Analysis of Source Bias}
The ``source bias" refers to the model's tendency to favor the source data, even when it is trained on both the source and target data. This occurs because the source data provides stronger supervision signals, leading to an imbalance in the training process. Empirical experiments have confirmed the existence of ``source bias" issue.

1) The presence of source bias is evident in the performance gap between the source and target data. For example, in Table~\ref{table:bias}, the \textit{Baseline} detector shows significantly better performance on the source data (48.91 vs. 39.56), highlighting a clear source bias.

2) Reduced bias is linked to better performance. In our experiments with Cityscapes$\rightarrow$Foggy Cityscapes, the latter includes three levels of fog density (0.02, 0.01, 0.005), with lower values indicating thinner fog. As shown in Table~\ref{table:bias1}, a closer resemblance to Cityscapes results in increasingly similar performance. (In the paper, fog level 0.02 is default as target domain). Addressing bias is possible, our proposed method effectively improves the performance of the source and target datasets.

3) We also conduct experiments on the object level. Firstly, Baseline and our detector are applied to the target images, followed by an assessment of the style similarity score of the detected boxes to the source-style. The style similarity score is measured using the trained instance-level domain discriminator. Then we evaluate the performance of the detectors across different similarity intervals. Table~\ref{table:bias2} results demonstrate that much higher performance is achieved when the detected objects are more similar to the source-style.


\begin{table}[t]
\caption{Detection performance on the different test set with various fog density.}
\small
\centering
\resizebox{0.47\textwidth}{!}{%
\begin{tabular}{c|c|c}
\hline
Test Set   & Baseline ($AP_{50}$) & Baseline +DSD ($AP_{50}$)  \\
\hline \hline
Foggy Cityscapes (0.02, Target)& 40.90 & 43.05 \\
\hline
Foggy Cityscapes (0.01) & 44.04 & 45.82 \\
\hline
 Foggy Cityscapes (0.005) &  46.83  &  47.56  \\
\hline
Cityscapes (Source) &  50.21  &  51.48  \\
\hline
\end{tabular}
}
\label{table:bias1}
\vspace{-0.5cm}
\end{table}
\begin{table}[t]
\caption{Detection performance in different similarity intervals at the object-level.}
\label{Ablation2}
\small
\centering
\resizebox{0.46\textwidth}{!}{%
\begin{tabular}{c|c|c|c|c}
\hline
Similarity Score   & 0-0.6	 & 0.6-0.75 & 0.75-0.85 & 0.85-1.0  \\
\hline \hline
Baseline ($AP_{50}$) &  5.7  &  13.6 &  37.9  & 65.0  \\
\hline
Baseline +DSD ($AP_{50}$) &  11.3  &  15.5 &  41.6  & 67.7  \\
\hline
\end{tabular}
}
\vspace{-0.5cm}
\label{table:bias2}
\end{table}

\subsection{Ablation Study}
In this section, we conduct ablation studies to validate our contributions. All experiments are conducted on the \textbf{\textit{FoggyCityscapes}} validation set with the DA-Faster as base detector. 

\textbf{Effectiveness of individual component}. We first investigate the impact of DSD and DCE on detection performance. As shown in Table~\ref{table:dua_LCR}, both DSD and DCE can improve the performance of the baseline under different backbone configurations. 
Finally, with all these components, we observe a respective enhancement in mAP of the baseline by 4.65\%, 5.07\%, and 4.64\% when employing ResNet50, ResNet101, and VGG16 as backbones. 
This demonstrates the effectiveness and necessity of DSD and DCE.

\textbf{The enhanced generalization of the DSD}.
In addition, to verify the improvement in detector's generalization brought by DSD, we evaluate the performance of the three methods on the source and target domain. As shown in Table~\ref{table:bias}, compared with the \textit{Source Only}, \textit{Baseline} and \textit{Baseline+DSD} improve the $AP_t$ by large margins, which demonstrates the positive impact of feature alignment. 
Furthermore, in comparison to \textit{Source Only}, \textit{Baseline} demonstrates a substantial enhancement exclusively in $AP_t$, whereas \textit{Baseline+DSD} exhibits significant improvements in both $AP_s$ and $AP_t$. This may be because our DSD framework distills more domain-agnostic knowledge to the detector, improving the generalization of detector on both source and target.

\textbf{Source or Target.}
In the DSD framework, distillation loss is utilized to guide the detector to learn domain-agnostic knowledge. 
We conduct an ablation study on choices of distillation data.
Since the annotations of target data are not available, as shown in Table~\ref{table:distillation}, we adopt three filtering strategies based on objectness, classification scores and objectness\&classification to select high-quality samples from RPN for distillation, respectively. The results indicate that when target data is involved in the distillation process, the model's performance decreases. This suggests that regardless of how we filter target samples, the selected samples inevitably contain a significant amount of noise, leading to a performance drop. In the end, we choose to distill only source samples in the DSD framework as the final solution, achieving the highest performance.

\begin{table}[]
 \vspace{-0.1cm}
\caption{Ablation study on the proposed DSD and DCE.}
\centering
\begin{tabular}{cc|ccc}
\hline
\multicolumn{2}{c|}{Module} & \multicolumn{3}{c}{mAP} \\ 

DSD          & DCE          & VGG16 & ResNet50    & ReNet101        \\ 
\hline \hline
             &             & 39.56  & 40.15       & 41.09          \\ 
\checkmark            &       & 42.00       & 43.05       & 42.63           \\ 
             & \checkmark          & 42.14   & 43.23       & 43.11        \\ 
\checkmark           & \checkmark       & \textbf{44.21}     & \textbf{45.22}       & \textbf{45.73}         \\
\hline
\end{tabular}

\label{table:dua_LCR}
 \vspace{-0.2cm}
\end{table}

\begin{table}[t]
\caption{Effects of distillation data for assessing DSD module in our method. $S$ means using source data for distillation. ${T}_{obj}$ and ${T}_{cls}$ represent filtering strategies based on objectness and classification, respectively, and ${T}_{obj\&cls}$ represents a strategy that filters both objectness and classification simultaneously. Here, the filtering threshold is set to 0.8, retaining samples with scores higher than 0.8.}
\centering
\begin{tabular}{c|ccc|c}
\hline
$S$                         & ${T}_{obj}$            & ${T}_{cls}$           & ${T}_{obj\&cls}$        & ${AP}_{50}$ \\ \hline \hline
                          &                           &                           &                           & 40.15          \\ 
\checkmark  &                           &                           &                           & \textbf{43.05}          \\ 
\checkmark  & \checkmark  &                           &                           & 38.87          \\ 
\checkmark  &                           & \checkmark  &                           & 37.64          \\ 
\checkmark  &                           &                           & \checkmark  & 38.43          \\ \hline
\end{tabular}

\label{table:distillation}
\vspace{-0.5cm}
\end{table}

\begin{figure}[t]
\centering
\includegraphics[width=1.00\columnwidth]{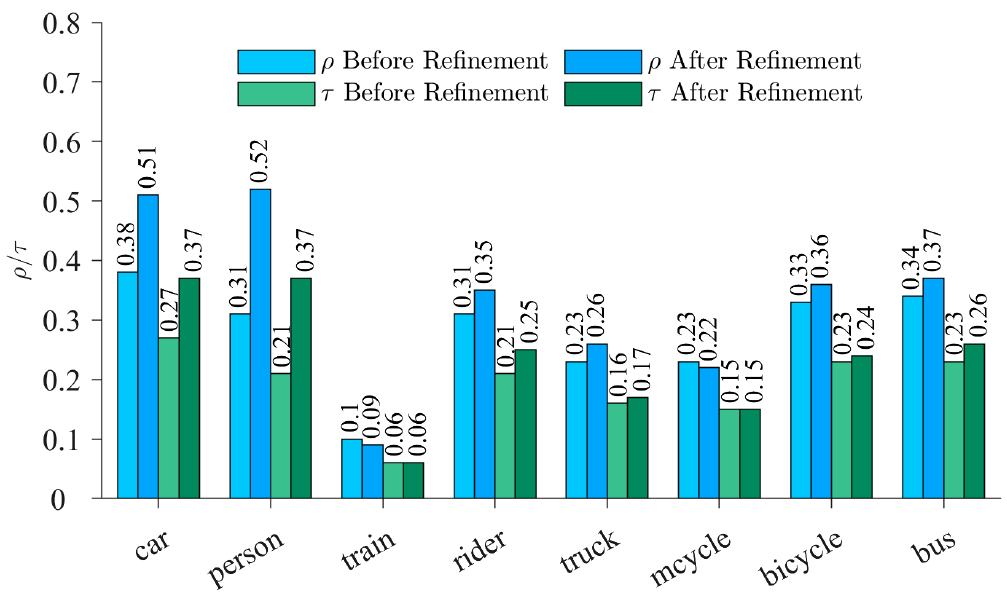} 
 \vspace{-0.4cm}
\caption{The variety of consistency across different classes before and after classification scores refinement on the \textbf{\textit{FoggyCityscapes}} test datasets.}
\label{fig8}
\end{figure}

\textbf{Choice of localization representation}. For brevity, we refer to centerness and IoU as $c$ and $b$. Here, we attempt to conduct ablation experiments on the localization-teacher and the localization representation.
As shown in Table~\ref{table:LCR}, when we train original OLN, using either c, b, or $\sqrt{{b}{c}}$ as localization representation to refine classification scores can improve the model's performance to different extents. This indicates that enhancing the consistency between classification and localization can effectively improve the detector's performance.
Furthermore, when training with TROLN (ours), using the localization score ($\sqrt{4{b}{c}\tau_{1}\tau_{2}}$) as the localization representation to calibrate classification scores, the model achieves the highest performance improvement (+2.17\% compared with baseline). This further validates the necessity and effectiveness of the TROLN training strategy and the DCE testing strategy. 
Moreover, we evaluate the variety of consistency in different classes before and after classification scores refinement on the test dataset. As shown in Fig.~\ref{fig8}, it can be observed that the consistency has been improved to varying degrees across almost all categories, further demonstrating the effectiveness of the DCE mechanism.

\begin{table}[]
\vspace{-0.4cm}
\caption{The effect of localization-teacher and localization representation.}
\centering
 \resizebox{0.48\textwidth}{!}{%
\begin{tabular}{c|llll|c|c}
\hline
\multirow{2}{*}{Training Stage} & \multicolumn{4}{l|}{DCE Testing Stage} & \multicolumn{1}{c|}{\multirow{2}{*}{Representation}} & \multicolumn{1}{l}{\multirow{2}{*}{$AP_{50}$}} \\ \cline{2-5}
                                & $c$       & $b$       & ${\tau}_{1}$       & ${\tau}_{2}$      & \multicolumn{1}{l|}{}                        & \multicolumn{1}{l}{}                      \\ \hline\hline
\multicolumn{1}{l|}{}           &         &         &          &         & \multicolumn{1}{l|}{}                        & 43.05                                     \\ \hline
\multirow{3}{*}{OLN}            & \checkmark       &         &          &         & $c$                                            & 43.95                                     \\
                                &         & \checkmark       &          &         & $b$                                            & 43.98                                     \\
                                & \checkmark       & \checkmark       &          &         & $\sqrt{{b}{c}}$                                           & 44.12                                     \\ \hline
\multirow{2}{*}{TROLN}          & \checkmark       & \checkmark       &          &         & $\sqrt{{b}{c}}$                                           & 44.76                                     \\
                                & \checkmark       & \checkmark       & \checkmark        & \checkmark       & $\sqrt{4{b}{c}\tau_{1}\tau_{2}}$                                       & \textbf{45.22}                                     \\ \hline
\end{tabular}
}
\label{table:LCR}
\vspace{-0.3cm}
\end{table}

\section{Conclusion}

To address source bias issue in domain adaptive object detection, we propose a distillation-based source debiasing framework. We train an instance-level classification-teacher model to guide the detector to acquire more domain-agnostic knowledge, improving the generalization on both domains. We also design a novel localization representation to refine classification scores, further improving the performance of the detector. Finally, our method achieved considerable improvement on several benchmark datasets under different base detectors for domain adaptation, demonstrating the effectiveness.

\section*{Acknowledgements}
This work was supported by Zhejiang Provincial Natural Science Foundation of China under Grant LD24F020016, and the National Natural Science Foundation of China under Grant 62176017.

\section*{Impact Statement}

This paper presents work whose goal is to advance the field of Deep Learning. There are many potential societal consequences of our work, none which we feel must be specifically highlighted here.

\bibliography{example_paper}
\bibliographystyle{icml2024}

\newpage
\appendix
\onecolumn
\section{More Implementation Details}
\label{sec:A}
\subsection{Classification-Teacher}
The classification-teacher model employs a ResNet101 \cite{he2016deep} architecture pre-trained on ImageNet \cite{russakovsky2015imagenet} as its backbone, with an input size of $224\times224$. Data augmentation strategies encompass random horizontal flipping, color distortion, Gaussian blur, and solarization. The AdamW optimizer is employed for optimizing the classification-teacher model, initialized with a learning rate of 0.0001 for 12 epochs. We decay the learning rate by ratio 0.1 at epoch 9 and 11 and the total batch size is set to 64. 
\subsection{Localization-Teacher (TROLN)}
\textbf{TROLN.} We reconstruct TROLN via adding a pixel-level global discriminator base on OLN \cite{kim2022learning}. As shown in Table~\ref{tab:more_arch_details}, we present the detailed architecture of this discriminator which consists of a Gradient Reversal Layer (GRL) and 4 conv layers. TROLN is trained with SGD optimizer with a 0.005 learning rate, 2 batch size for 12 epochs and we decay the learning rate by ratio 0.1 at epoch 6 and 7. 
	\begin{table}[h]
        \caption{\label{tab:more_arch_details}Architectures of the adversarial alignment modules.}
		\small
		\begin{center}
			\begin{tabular}{lll}
				\hline
				\multicolumn{3}{c}{Global Discriminator} \\
				\hline \hline
				\multicolumn{3}{c}{Gradient Reversal Layer (GRL)} \\
				\multicolumn{3}{c}{Conv 256 $\times$ 3 $\times$ 3, stride 1 $\to$  LeakyReLU slope 0.2} \\
				\multicolumn{3}{c}{Conv 128 $\times$ 3 $\times$ 3, stride 1 $\to$  LeakyReLU slope 0.2} \\
				\multicolumn{3}{c}{Conv 128 $\times$ 3 $\times$ 3, stride 1 $\to$  LeakyReLU slope 0.2} \\
				\multicolumn{3}{c}{Conv 1 $\times$ 3 $\times$ 3, stride 1}\\
				\hline
			\end{tabular}
		\end{center}
		
	\end{table}


\subsection{DSD Training Based on AT and CMT}
Due to original AT \cite{li2022cross} or CMT \cite{cao2023contrastive} uses two-stage training, here we train AT or CMT for 20k iteration in the first stage and 10k iterations in the second stage. Moreover, our DSD module is only added to the detector in the second stage. Other hyper-parameters are the same as in the original implementation of AT and CMT. Our implementation is based on Detectron2 and the publicly available code by AT and CMT. Each experiment is conducted on 4 NVIDIA 3090 GPUs.
\section{Further Ablation Studies}
\label{sec:B}
To further analyze the effect of the data on the DSD and TROLN, we conduct extensive ablation studies in this section. Here, all experiments are done with DA-Faster \cite{chen2018domain} with ResNet50 backbone on \textit{FoggyCityscapes} test set.

\subsection{IoU Threshold in DSD Framework}

We empirically choose IoU threshold $T$ to analyze how $T$ affects the detector's performance in the DSD framework.
As shown in Table~\ref{Ablation2}, we test a range of $T$. On one hand, when $T$ is smaller, the filtered positive samples have significant differences from the ground truth, introducing extra noise into the classification-teacher model and causing a drop in model performance. On the other hand, a larger $T$ results in a significant reduction in the number of samples, weakening the effect of distillation. Eventually, we set $T=0.8$ with the best performance.
\begin{table}[h]
\caption{Effects of $T$ for estimating the DSD module.}
\label{Ablation2}
\small
\centering
\begin{tabular}{c|c|c|c|c|c|c}
\hline
$T$   & 0.65 & 0.7 & 0.75 & 0.8 & 0.85 & 0.9 \\
\hline \hline
mAP &  40.12  &  40.64 &  41.86  &\textbf{43.05}   &42.56 &  42.12   \\
\hline
\end{tabular}
\end{table}

\subsection{Choices of Training Data for TROLN}
In order to investigate the effect of training data on TROLN, we conduct an ablation study on choices of training data. As shown in Table~\ref{table:DEC}, we train the original OLN \cite{kim2022learning} using source data ($S$), source2target data ($T'$) and $S$\&$T'$, respectively. 
It's worth noting that training OLN solely on $T'$ data can not effectively mine target related information. This may be because there is still a domain shift between generation data $T'$ and the real target data. Moreover, training OLN based on both $S$ and $T'$ effectively enhances baseline performance. This indicates that mixed-style data is beneficial for the localization network to generalize to target data.
Finally, when incorporating the domain affinity weights $\tau_{1}$ and $\tau_{2}$ into both the training of TROLN and the design of localization quality representation, the model achieves the highest performance (45.22\%).
\begin{table}[h]
\setlength{\tabcolsep}{3pt}
\caption{Effects of Training data in TROLN on detector's performance.}
\small
\centering
\begin{tabular}{l|cc|cccc|c|c}
\hline
\multicolumn{1}{c|}{\multirow{2}{*}{Method}} & \multicolumn{2}{c|}{Training Data}           & \multicolumn{4}{c|}{DCE Testing Stage}                                                         & \multirow{2}{*}{Metric} & \multirow{2}{*}{$AP_{50}$} \\ \cline{2-7}
\multicolumn{1}{c|}{}                        &  \multicolumn{1}{p{0.7cm}<{\centering}}{$S$}                    &  \multicolumn{1}{p{0.7cm}<{\centering}|}{$T'$}                    & \multicolumn{1}{p{0.4cm}<{\centering}}{$c$}                    & \multicolumn{1}{p{0.4cm}<{\centering}}{$b$}                    & \multicolumn{1}{p{0.4cm}<{\centering}}{${\tau}_{1}$} & \multicolumn{1}{p{0.4cm}<{\centering}|}{${\tau}_{2}$} &                         &                       \\ \hline\hline
                                             & \multicolumn{1}{l}{} & \multicolumn{1}{l|}{} & \multicolumn{1}{l}{} & \multicolumn{1}{l}{} &                        &                         & \multicolumn{1}{l|}{}   & 43.05                 \\ \hline
\multirow{3}{*}{OLN}                         & \checkmark                    & \multicolumn{1}{l|}{} & \checkmark                    & \checkmark                    &                        &                         & $\sqrt{{b}{c}}$                      & 35.76                 \\ 
                                             & \multicolumn{1}{l}{} & \checkmark                     & \checkmark                    & \checkmark                    &                        &                         & $\sqrt{{b}{c}}$                      & 41.96                 \\ 
                                             & \checkmark                    & \checkmark                     & \checkmark                    & \checkmark                    &                        &                         & $\sqrt{{b}{c}}$                      & 44.12                 \\ \hline
\multicolumn{1}{c|}{\multirow{2}{*}{TROLN}}  & \checkmark                    & \checkmark                     & \checkmark                    & \checkmark                    &                        &                         & $\sqrt{{b}{c}}$                      & 44.76                 \\ 
\multicolumn{1}{c|}{}                        & \checkmark                    & \checkmark                     & \checkmark                    & \checkmark                    & \multicolumn{1}{c}{\checkmark}  & \multicolumn{1}{c|}{\checkmark}  & $\sqrt{4{b}{c}\tau_{1}\tau_{2}}$                  & \textbf{45.22}                 \\ \hline
\end{tabular}
\label{table:DEC}
\end{table}

\begin{figure}
   \centering
   \includegraphics[width=0.70\linewidth]{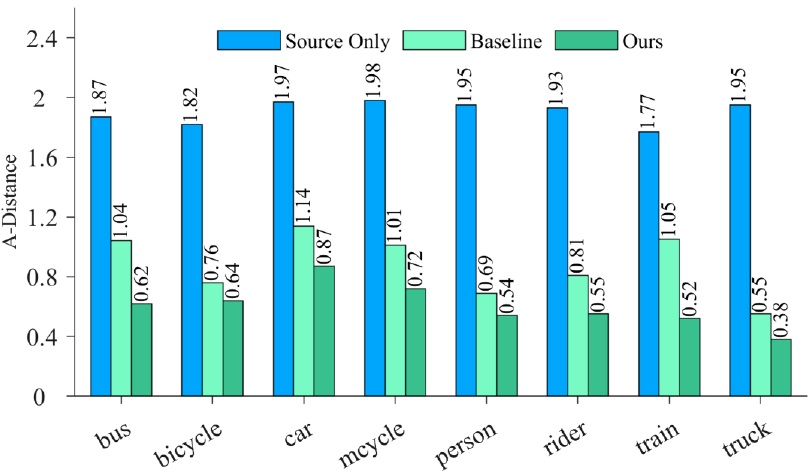}
   \caption{Feature distribution discrepancy of foregrounds.}
   \label{fig:distance} 
\end{figure}

\begin{figure}[t]
\centering
\includegraphics[width=0.70\columnwidth]{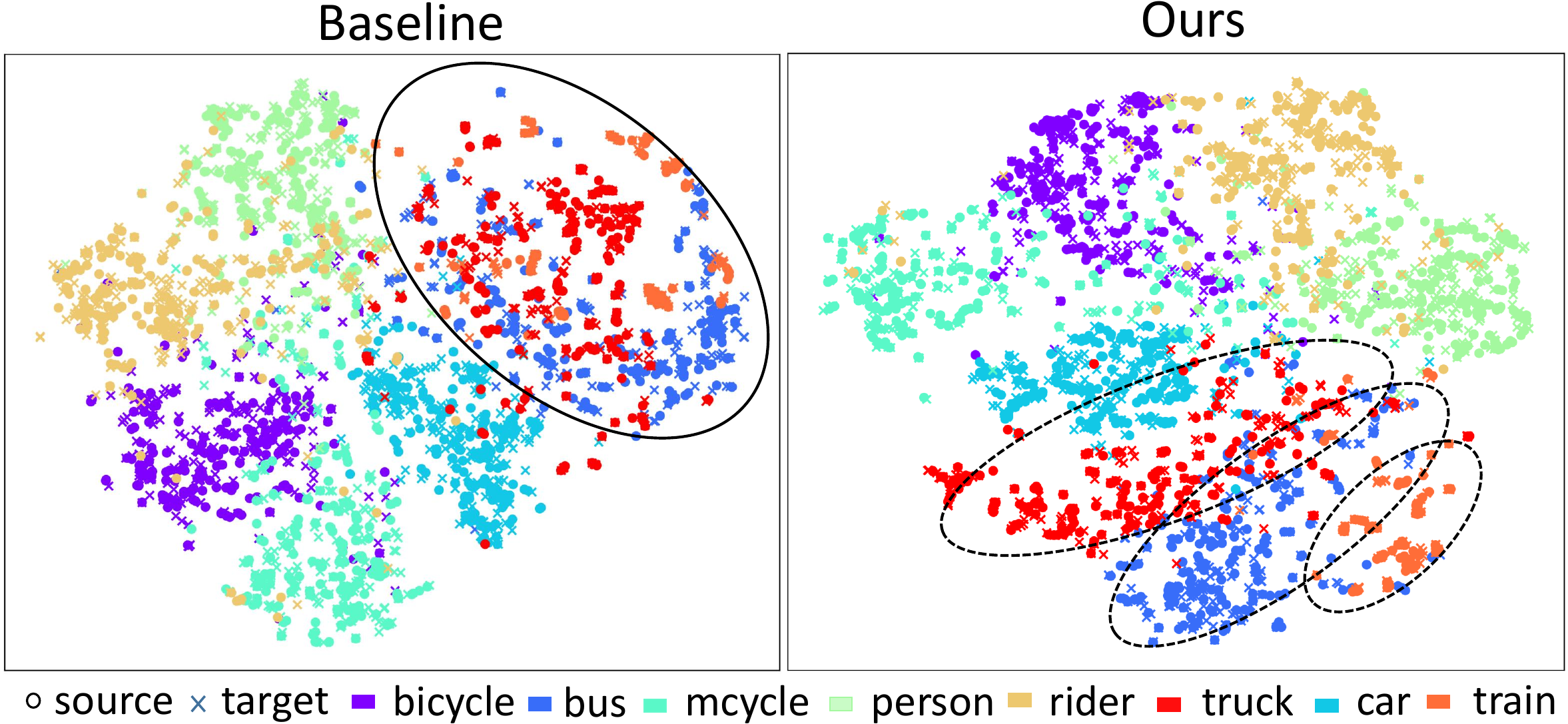} 
\caption{T-SNE visualization of features produced by the baseline model and our method.}
\label{fig9}
\end{figure}


\begin{table}[t]
\caption{TIDE error analysis. The $\Delta$$AP^{box}\texttt{@}0.5$ metric is defined as how much $AP_{50}$ can be added to the detector if an oracle fixes a certain error type in TIDE~\cite{bolya2020tide}.}
\centering
\begin{tabular}{p{0.6cm}<{\centering} p{0.6cm}<{\centering}|p{0.6cm}<{\centering}p{0.6cm}<{\centering}p{0.6cm}<{\centering}p{0.6cm}<{\centering}p{0.6cm}<{\centering}p{0.6cm}<{\centering}}
\hline
\multicolumn{2}{c|}{Module} & \multicolumn{6}{c}{$\Delta$$AP^{box}\texttt{@}0.5$}                                                                        \\ \hline \hline
DSD          & DCE          & \textit{cls}$\downarrow$           & \textit{loc}$\downarrow$           & \textit{both}$\downarrow$          & \textit{dup}$\downarrow$           & \textit{bg}$\downarrow$            & \textit{miss}$\downarrow$           \\ \hline
             &              & 5.45          & 12.46         & 1.41          & 0.01          & \textbf{1.32} & 15.62          \\
\checkmark            &              & 4.94          & 9.95          & \textbf{1.40} & \textbf{0.00} & 1.47          & 17.07          \\
\checkmark            & \checkmark            & \textbf{3.78} & \textbf{9.06} & 1.58          & 0.05          & 1.97          & \textbf{13.79} \\ \hline
\end{tabular}

\label{table:error}
\end{table}

\begin{figure*}[t]
	\centering
        \includegraphics[width=1.00\textwidth]{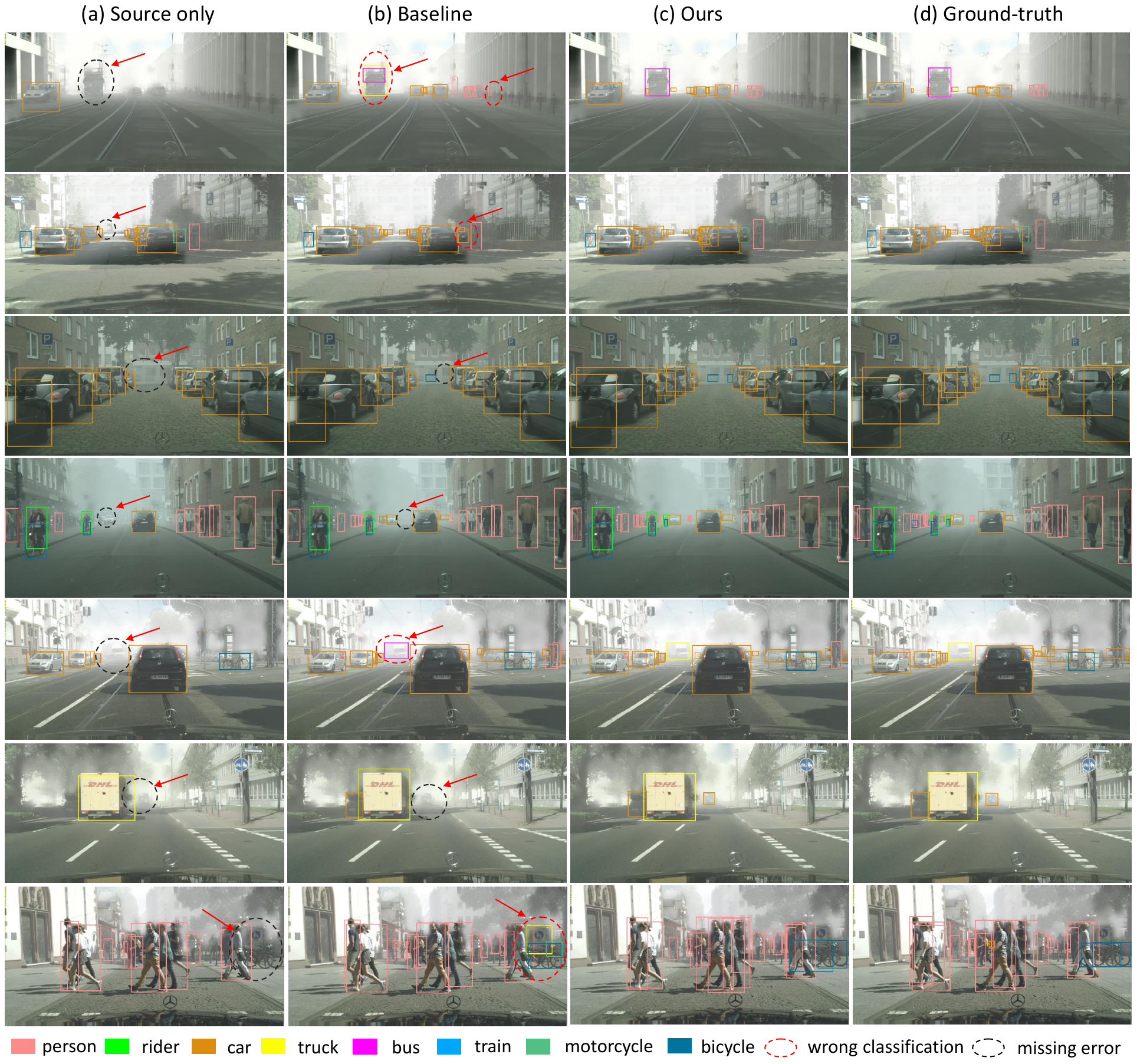}
	\caption{Qualitative results on the \textbf{\textit{Cityscapes-to-FoggyCityscapes}} adaptation scenario of (a) the \textit{Source Only} model, (b) \textit{Baseline} \cite{chen2018domain}, (c) \textit{Ours}, and (d) \textit{Ground-truth}. (Zooming in for best view.)}
 \label{fig:results}
\end{figure*}

\section{Further Analysis}
\label{sec:C}

\subsection{Distribution Discrepancy of Foregrounds}
 The theoretical result in~\cite{ben2006analysis} indicates that $\mathcal{A}$-distance can serve as a metric for quantifying domain discrepancy. In practice, we calculate the Proxy $\mathcal{A}$-distance as an approximation, which is defined as $d_{\mathcal{A}} = 2(1 - \epsilon)$. Here, $\epsilon$ represents the generalization error of a binary classifier (implemented with two fully connected layers in our experiments) that tries to distinguish which domain the ROI features come from. Fig.~\ref{fig:distance} displays the distances for each category on the \textbf{\textit{Cityscapes-to-FoggyCityscapes}} task with the foreground features extracted from the models of \textit{Source Only} \cite{ren2015faster}, \textit{Baseline} \cite{chen2018domain} and \textit{Ours}. Compared with the \textit{Source Only} model, \textit{Baseline} and \textit{Ours} reduce the distances in all the categories by large margins, which demonstrates the necessity of feature alignment. Furthermore, since we utilize a classification-teacher to distill domain-agnostic knowledge to the detector, we achieve a smaller $\mathcal{A}$-distance compared to \textit{Baseline}.

 After training, for each category, we randomly sample the same number of ROI features from the source and target domain for T-SNE visualization, as shown in Fig.~\ref{fig9}. It can be observed that those similar categories (truck, bus and train) can be separated clearly by our method, which benefits the following detection.

 \subsection{Error Analysis of Detection Results} 

To further validate the effect of the proposed framework for cross-domain object detection, we analyze the detection errors of the models of \textit{Baseline}, \textit{Baseline+DSD} and \textit{Baseline+DSD+DCE (Ours)}  via the TIDE toolbox~\cite{bolya2020tide} on the \textbf{\textit{Cityscapes-to-FoggyCityscapes}} task. As shown in Table~\ref{table:error}, we follow TIDE to categorize the detection errors into six types: \emph{cls}: localized correctly, but classified incorrectly; \emph{loc}: classified correctly, but localized incorrectly; \emph{both}: classified incorrectly and localized incorrectly; \emph{dup}: two or more detected boxes match with the same ground-truth box; \emph{bg}: classifying the background as foreground mistakenly; \emph{miss}: foreground objects are not detected by the detector. (See~\cite{bolya2020tide} for more details and discussion.)

We observe that both \textit{Baseline+DSD} and \textit{Ours} make fewer classification and localization errors than \textit{Baseline}. It indicates that the DSD module effectively distills domain-agnostic features from the classification-teacher to the detector, guiding the detector to extract the superior feature representation of the foreground.
Besides, \textit{Ours} performs the best in terms of the \emph{miss} error category among the three models. This further illustrates that employing DCE to refine the classification scores can effectively improve the consistency between classification and localization. The DCE enables detected boxes with better localization to also have higher classification scores, thereby reducing \emph{miss} error.

\section{Qualitative Results}
\label{sec:D}

We present more qualitative comparisons among (a) \textit{Source Only}, (b) \textit{Baseline} \cite{chen2018domain}, (c) \textit{Ours}, and (d) \textit{Ground-truth} in Fig.~\ref{fig:results}. Our method can eliminate some missing errors and avoid some wrong classification cases compared with the \textit{Baseline}, which verifies the effectiveness of proposed 
DSD module and DCE strategy.
\end{document}